%% file: main.tex
\documentclass[conference]{IEEEtran}
\IEEEoverridecommandlockouts

\usepackage{cite}
\usepackage{amsmath,amssymb,amsfonts}
\usepackage{algorithmic}
\usepackage{graphicx}
\usepackage{textcomp}
\usepackage{xcolor}
\usepackage[numbers]{natbib}
\usepackage{listings}
\usepackage{xcolor}
\usepackage{tcolorbox}

\usepackage{comment}
\usepackage{tikz}
\usetikzlibrary{arrows.meta, automata, positioning}
\lstdefinelanguage{json}{
    basicstyle=\ttfamily,
    comment=[l]{//},
    morestring=[b]",
    moredelim=[s][\color{black}]{:}{\ },
    stringstyle=\color{blue},
    literate=
     *{0}{{{\color{purple}0}}}{1}
      {1}{{{\color{purple}1}}}{1}
      {2}{{{\color{purple}2}}}{1}
      {3}{{{\color{purple}3}}}{1}
      {4}{{{\color{purple}4}}}{1}
      {5}{{{\color{purple}5}}}{1}
      {6}{{{\color{purple}6}}}{1}
      {7}{{{\color{purple}7}}}{1}
      {8}{{{\color{purple}8}}}{1}
      {9}{{{\color{purple}9}}}{1}
      {:}{{{\color{black}{:}}}}{1}
      {,}{{{\color{black}{,}}}}{1}
}
\def\BibTeX{{\rm B\kern-.05em{\sc i\kern-.025em b}\kern-.08em
    T\kern-.1667em\lower.7ex\hbox{E}\kern-.125emX}}
\newboolean{anonymized}
\setboolean{anonymized}{false}

\begin{document}

\title{Can Large Language Models Help Developers with Robotic Finite State Machine Modification?\ifthenelse{\boolean{anonymized}}{\thanks{Funding acknowledgements omitted for anonymization.}}{}}
\ifthenelse{\boolean{anonymized}}{
    \author{
        \IEEEauthorblockN{Anonymous}
        \IEEEauthorblockA{
            Anonymous \\
            Anonymous, Anonymous \\
            Anonymous
        }
    }
}{
    \author{
        \IEEEauthorblockN{
            Xiangyu Robin Gan\IEEEauthorrefmark{1},
            Yuxin Ray Song\IEEEauthorrefmark{1},
            Nick Walker\IEEEauthorrefmark{1},
            Maya Cakmak\IEEEauthorrefmark{1} 
        }
        \IEEEauthorblockA{
            \IEEEauthorrefmark{1}Paul G. Allen School of Computer Science \& Engineering, University of Washington, Seattle, USA\\
            Emails: \{robingan, syx1995, nswalker, mcakmak\}@cs.washington.edu
        }
    }
}

\maketitle

\begin{abstract}
\input{sections/0_Abstract}
\end{abstract}


\input{sections/1_Introduction.tex}

\input{sections/2_Related_Works.tex}
\input{sections/3_FSM_Manipulation}
\input{sections/4_Evaluation.tex}
\input{sections/5_Results_Discussion}
\input{sections/6_Conclusion}
\input{sections/7_Acknowledgement}

\newpage
\bibliographystyle{IEEEtranN}
\bibliography{references}

\newpage
\appendices

\end{document}

%% file: sections/0_Abstract.tex
Finite state machines (FSMs) are widely used to manage robot behavior logic, particularly in real-world applications that require a high degree of reliability and structure. However, traditional manual FSM design and modification processes can be time-consuming and error-prone. We propose that large language models (LLMs) can assist developers in editing FSM code for real-world robotic use cases. LLMs, with their ability to use context and process natural language, offer a solution for FSM modification with high correctness, allowing developers to update complex control logic through natural language instructions. Our approach leverages few-shot prompting and language-guided code generation to reduce the amount of time it takes to edit an FSM. To validate this approach, we evaluate it on a real-world robotics dataset, demonstrating its effectiveness in practical scenarios.

%% file: sections/1_Introduction.tex
\section{Introduction}
Finite State Machines (FSMs) play a critical role in various domains such as robotics, software engineering, and control systems, where they model sequential logic and event-driven behaviors~\cite{Boren2010TheSH}. The benefits of using FSMs include their clear structure, deterministic nature, and ability to effectively model complex systems with distinct states and transitions, providing an intuitive framework for designing reactive behaviors and ensuring predictable system performance. In robotics, FSMs serve as essential tools for managing stateful logic, enabling robots to transition smoothly between behaviors based on sensor input, task requirements, or environmental changes. For example, in robotic manipulation tasks, an FSM can dictate the sequence of grasping, lifting, and placing objects, where each action depends on the successful completion of the previous state.

As the complexity of robotic systems increases, designing and manipulating FSMs becomes a daunting challenge due to the sheer number of states, transitions, and conditions to be considered. For instance, a household service robot participating in the RoboCup@Home competition may need to handle multiple tasks like object recognition, navigation, and user interaction, each requiring intricate behavior logic and dynamic responses to real-world stimuli. This complexity necessitates tools that can aid robot developers in automating FSM design, modification, and analysis, reducing the cognitive and developmental overhead involved in manually editing state machine logic.

With the rapid advancement of Large Language Models (LLMs) in recent years, models like GPT-4 have exhibited remarkable capabilities in natural language understanding, code generation, and problem-solving across diverse fields. LLMs have been applied to various tasks in the robotics domain, including but not limited to vision-language planning (e.g., RT-2~\cite{brohan2023rt2visionlanguageactionmodelstransfer}), robot code generation ~\cite{codeaspolicies2022}, and semantic parsing for task execution ~\cite{liu2024okrobot}. Their ability to interpret and generate structured information suggests a potential to facilitate FSM manipulation by automating aspects of system design and enabling easier updates through natural language inputs.

\begin{figure}[t]
    \centering
    \includegraphics[width=9cm]{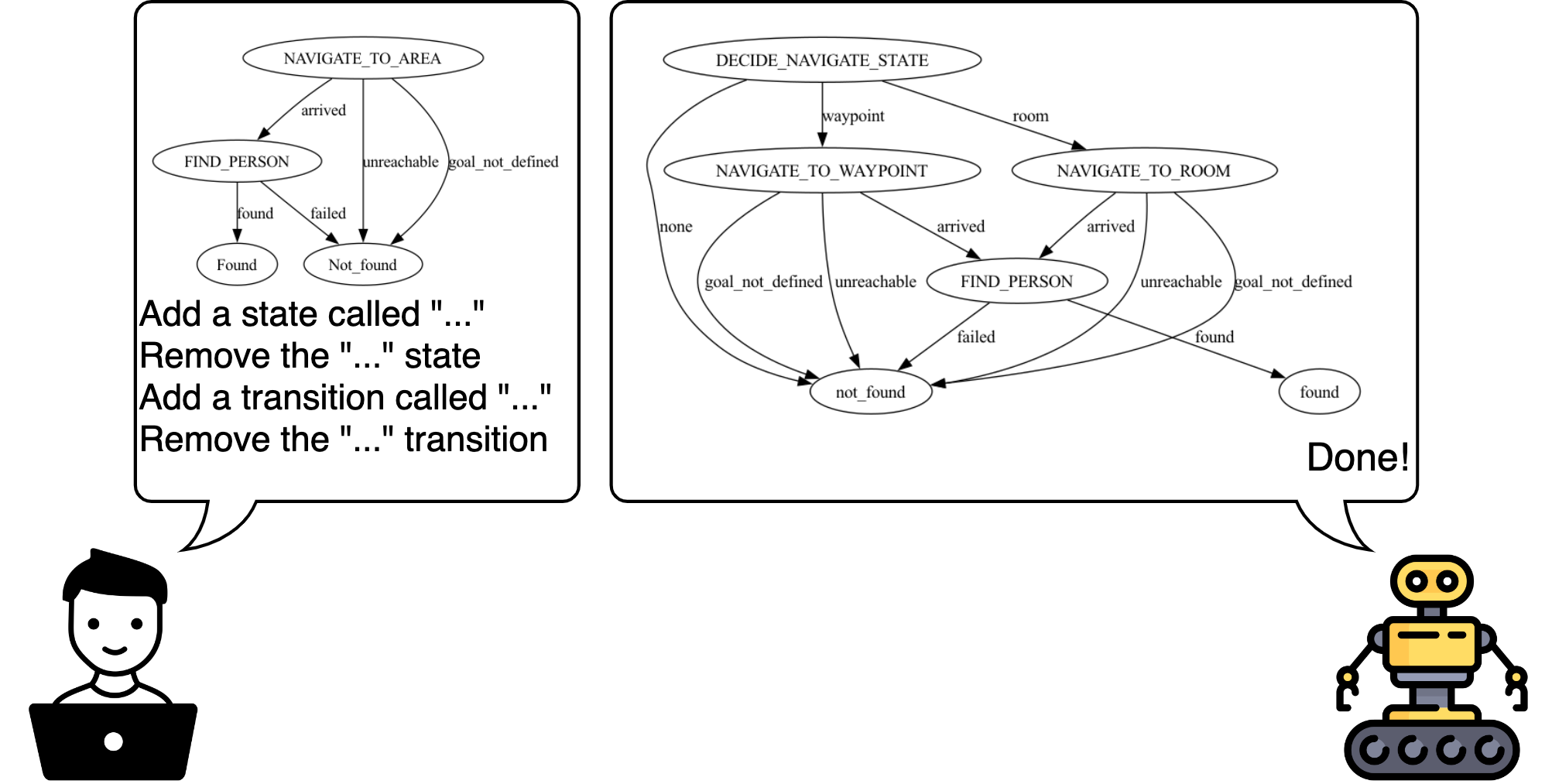}
    \caption{Robotics developers can use natural language to modify finite state machines through a chatbot.}
    \label{fig:your_image_label}
\end{figure}

In this paper, we investigate the question: can large language models help developers with robotic finite state machine modification? We explore the capabilities and limitations of LLMs in the context of FSM manipulation. We introduce ChatFSM, a tool that integrates LLM APIs with Retrieval-Augmented Generation (RAG) ~\cite{DBLP:journals/corr/abs-2005-11401} to support the manipulation and visualization of FSMs. To evaluate whether LLMs are well-suited to modifying realistic robot FSMs, we gather a dataset of state machine modifications from the version control history of the Eindhoven University of Technology's RoboCup@Home team. This dataset provides concrete examples of FSM modifications in a real-world robotic application. We assess the performance of ChatFSM by comparing its ability to handle FSM manipulation tasks against the equivalent tasks performed by human developers. Our findings reveal that ChatFSM successfully reproduces commit changes to FSM logic across multiple files within the repository, demonstrating its potential to assist in complex FSM editing tasks and automate behavior modifications efficiently.

By examining the strengths and weaknesses of LLM-based FSM manipulation, this paper contributes to the growing body of research on leveraging AI for automated system design and aims to highlight the potential role of LLMs in simplifying and accelerating robotic behavior development.

%% file: sections/2_Related_Works.tex
\section{Related Works}

\subsection{Finite State Machines}\label{AA} Finite State Machines (FSMs) are mathematical models used to represent systems with a finite number of states and transitions between those states triggered by events or conditions. FSMs are employed across a variety of domains, including digital circuit design~\cite{DBLP:journals/corr/abs-1003-0522}, language parsing~\cite{7724306}, software testing, and behavior specification in robotics~\cite{Boren2010TheSH}. In robotics, FSMs are often used to structure control flow, defining behavior sequences that are both robust and predictable. Behavior Trees (BTs) ~\cite{colledanchise2018behavior} have also been widely adopted in robotic behavior design as an alternative to FSMs due to their modular and hierarchical nature, allowing for more flexible task decomposition and reuse. However, FSMs still remain a popular choice because of their straightforward representation and the ease of defining transitions explicitly, which is crucial in systems requiring deterministic behavior control.

Recent works have explored hybrid approaches that combine the strengths of FSMs and other models like BTs and hierarchical FSMs (HFSMs) to allow for more scalable designs ~\cite{lee2024efficiency}. These developments reflect an increasing interest in finding representations that balance clarity, flexibility, and scalability, particularly in the context of complex robotic behaviors that must adapt to dynamic environments while ensuring a manageable structure for system developers.

\subsection{LLMs as High-level Planners}\label{AA} Recent works have demonstrated the potential of Large Language Models (LLMs) as high-level robotic planners, particularly in response to natural language instructions~\cite{han2024llmpersonalize, ahn2022icanisay, 10161317}. LLMs have been leveraged to translate user-provided goals or commands into sequences of actions that robots can perform in various domains, including manipulation, navigation, and interaction tasks. For example, studies have showcased how LLMs can dynamically generate action plans and help robots adapt to unforeseen circumstances by reinterpreting human inputs, facilitating on-the-fly adjustments and replanning~\cite{huang2022innermonologueembodiedreasoning, shi2024yellrobotimprovingonthefly}. This capability allows robots to execute high-level reasoning, enhancing their ability to operate autonomously or in close collaboration with humans.

While these advancements offer significant flexibility, they often lack an explicit and structured behavior representation, which can limit the ability to introduce case-specific modifications or debugging necessary for safe and reliable robotics applications. FSMs, on the other hand, offer a clear and explicit representation of robot behavior, enabling the introduction of detailed transitions and fail-safe mechanisms. However, their manual modification can be time-consuming and error-prone. Our work builds on this by exploring the ability of LLMs to modify FSMs explicitly and directly, demonstrating that language models can automate complex code generation and FSM transformations in response to high-level natural language inputs.

\subsection{LLMs and Code Generation for FSMs}\label{AA} Some studies have started exploring the use of LLMs in automating code modifications. Prior research has shown that LLMs, such as Codex and Copilot, can assist developers by generating code based on natural language descriptions, providing solutions to programming problems, and even conducting bug fixes~\cite{chen2021evaluating, mastropaolo2023robustness}. While these applications are not specifically focused on FSMs, they indicate the potential for LLMs to automate code generation in structured programming domains. Some work has begun to look into how LLMs might assist in generating FSMs directly from natural language descriptions or task specifications, although this remains an emerging area.

\subsection{Retrieval-Augmented Generation}\label{AA} Retrieval-Augmented Generation (RAG) is a technique that enhances language models by providing additional contextual information retrieved from a database or corpus, significantly improving the accuracy, coherence, and relevance of generated outputs~\cite{DBLP:journals/corr/abs-2005-11401}. RAG has proven effective in many natual language processing tasks, including question-answering, information retrieval, and even contextual code generation, where the retrieved information serves as a critical source of guidance for model outputs.

In our work, we integrate RAG into the FSM modification process by leveraging relevant code and documentation as additional context for the LLM to explore the potential of increasing the correctness of FSM modification.

\subsection{FSM Tools and Automation}\label{AA} Various tools have been developed to facilitate FSM design and manipulation, from graphical editors that allow developers to visually create and modify FSMs to libraries that automate FSM execution in robotics and software engineering. Tools such as ROS state machine libraries (such as SMACH ~\cite{smach} and FlexBE ~\cite{FlexBE}) provide structured environments to create FSMs in robot control architectures, and state machine frameworks like SCXML ~\cite{barnett2017introduction} support FSM implementation in software development. However, these tools often rely on manual coding, which can be cumbersome, particularly as FSM complexity increases.

Our research positions LLMs as an extension of these tools, providing a natural language interface for modifying FSMs, thus reducing manual effort and enabling more intuitive FSM management through natual language interactions.

%% file: sections/3_FSM_Manipulation.tex
\section{LLM-Based State Machine Modification}
\subsection{Overview}

To demonstrate that LLMs can modify FSMs based on user requests in natural language, we developed \textsc{ChatFSM}, an LLM-based agent specifically designed to handle FSM modifications(we define an agent as a LLM invocation).

We define an FSM as a labeled directed graph $G = (S, T)$ consisting of vertices $S$ and transitions $T$. Each vertex is an instance $S = \{S_0 ... S_n\}$. Each state implements a robot action and defines a set of conditions $O^{S_n} = \{O^{S_n}_0 ... O^{S_n}_n\}$.
Each vertex has a complete mapping of its outcomes to other vertices, i.e. one from-edge for each outcome in its respective outcome set $O^{S_n}$ mapping to another vertex. 
Execution begins in initial state $S^I$ and ends in sink states $S^X = \{ S^X_0... S^X_n\}$ which denote the outcome of the state machine.
Each state is assigned a unique label.

\begin{figure}[h!]
    \centering
    \resizebox{0.5\textwidth}{!}{%
    \begin{tikzpicture}[shorten >=1pt, node distance=2.5cm, on grid, auto]
        \node[state, initial] (S0) {$S^I$};
        \node[state] (S1) [right=of S0] {$S_1$};
        \node[state] (S2) [right=of S1] {$S_2$};
        \node[state, accepting] (S3) [right=of S2] {$S^X_0$};
        
        \path[->] 
        (S0) edge [bend left] node {$O^{S^I}_0$} (S1)
        (S1) edge [bend left] node {$O^{S_1}_0$} (S2)
        (S2) edge [bend left] node {$O^{S_2}_0$} (S3)
        (S1) edge [loop above] node {$O^{S_1}_1$} (S1)
        (S2) edge [bend right] node [below] {$O^{S_2}_1$} (S1);
    \end{tikzpicture}
    }
    \caption{A FSM as a directed graph $G = (S, T)$. The vertices $S$ represent states ($S^I, S_1, S_2, S^X_0$), with $S^I$ as the initial state and $S^X_0$ as a sink state indicating the FSM's termination. Each state has labeled transitions $T$ (e.g., $O^{S_n}_i$) mapping to outcomes leading to other states or looping within the same state, demonstrating how transitions operate within the FSM structure.}
    \label{fig:fsm_model}
\end{figure}
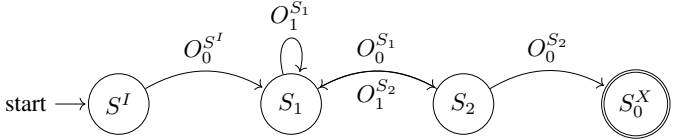

\begin{figure}[h!]
    \centering
    \resizebox{0.5\textwidth}{!}{%
    \begin{tikzpicture}[shorten >=1pt, node distance=3cm, on grid, auto]
        \node[state, initial] (S0) {Idle};
        \node[state] (S1) [right=of S0] {Navigate};
        \node[state] (S2) [right=of S1] {Open Door};
        \node[state] (S3) [right=of S2] {Enter Room};
        \node[state, accepting] (S4) [right=of S3] {Destination};
        
        \path[->] 
        (S0) edge [bend left] node {Start Command} (S1)
        (S1) edge [bend left] node {Reached Door} (S2)
        (S2) edge [bend left] node {Door Opened} (S3)
        (S3) edge [bend left] node {Room Entered} (S4)
        (S2) edge [loop below] node {Failed to Open Door} (S2)
        (S1) edge [loop below] node {Obstacle Detected} (S1);
    \end{tikzpicture}
    }
    \caption{A FSM representing a robot's navigation task. The states include "Start," "Navigate," "Open Door," "Enter Room," and "Destination." Transitions between these states are driven by specific conditions.}
    \label{fig:robot_navigation_fsm}
\end{figure}
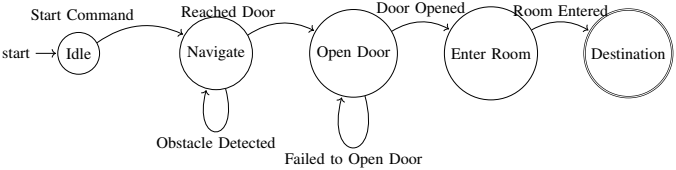

\subsection{ChatFSM}

\textbf{The agent is prompted with FSM code in a programming language, followed by a user message detailing the requested changes}. ChatFSM uses a system prompt which specializes its behavior to the task: 

\begin{quote}
You are an advanced language model specialized in finite state machines (FSMs). Users will provide you with two main inputs:

\item 1. \textbf{FSM Code}: This is a code representation of a finite state machine. The code will define states, transitions, inputs, and outputs.
\item 2. \textbf{Requested Changes}: These are specific modifications the user wants to make to the provided FSM code. Changes could include adding, removing, or altering states, transitions, inputs, or outputs.

Your task is to:

\item 1. \textbf{Understand the FSM Code}: Parse and comprehend the structure and logic of the provided FSM code.
\item 2. \textbf{Implement the Requested Changes}: Modify the FSM code according to the user's specified changes.
\item 3. \textbf{Return the Updated FSM Code}: Provide the user with the revised FSM code that reflects the requested modifications.

Ensure the following when processing requests:
- Maintain the integrity and functionality of the FSM.
- Validate the changes for logical consistency within the FSM structure.
- Provide clear and concise updated FSM code.

Respond with the entire code. Do not skip parts. Respond with the code only, nothing else. Finite State Machine (FSM) code follows the Requested Changes.
\end{quote}


The interface also includes a visualization module, which leverages an LLM-powered agent to extract the FSM from the input and convert it into a standard JSON format. This process, explained in detail in the evaluation section as part of the assessment steps, utilizes the Graphviz package to generate corresponding visual representations. This feature aids developers by facilitating easier analysis and debugging.

\begin{figure}[t]
    \centering
    \includegraphics[width=8cm]{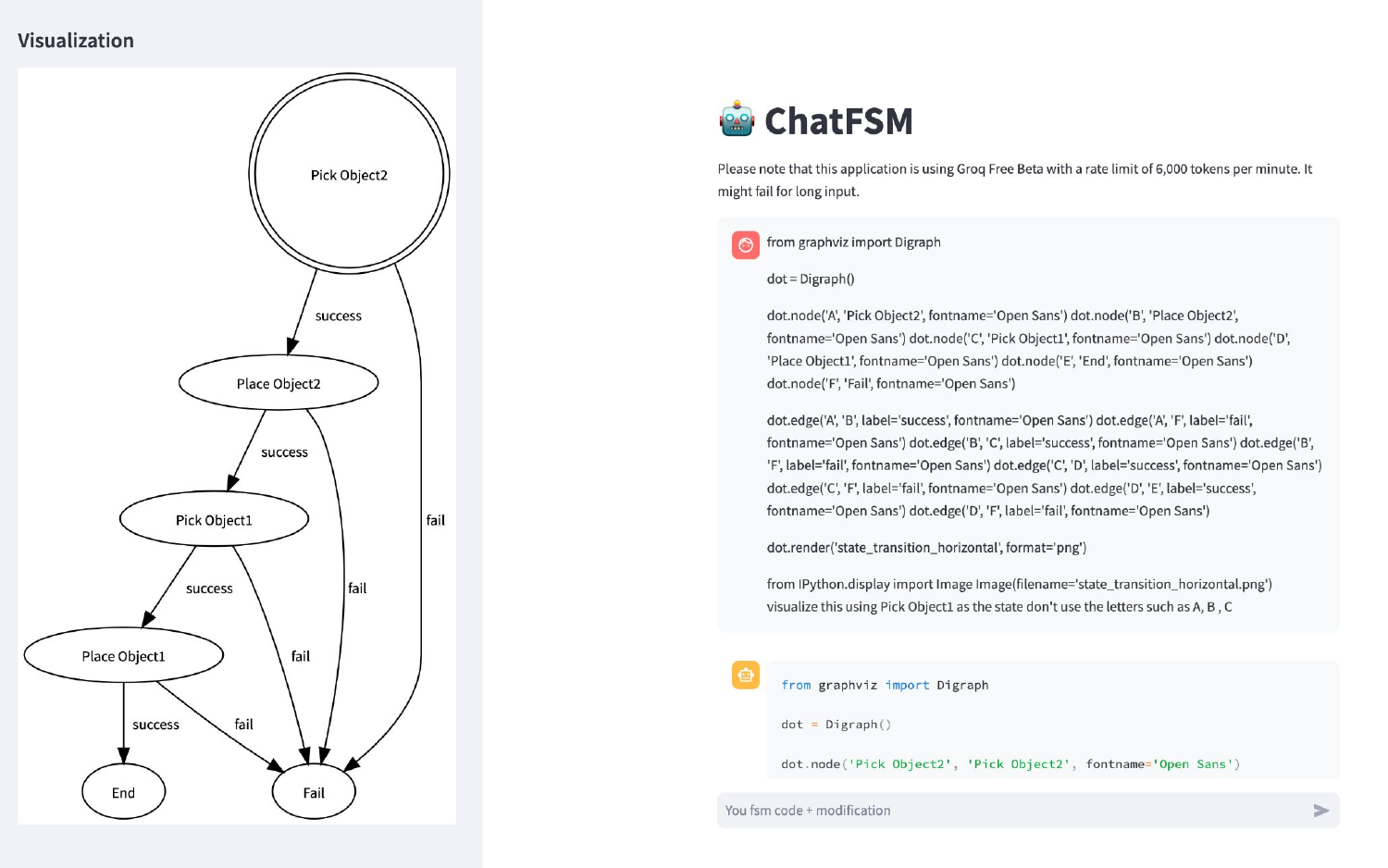}
    \caption{ChatFSM Interface, FSM Visualization is on the left, Chatbot input interface is on the right.}
\end{figure}

%% file: sections/4_Evaluation.tex
\section{Evaluation}

We evaluate how accurately a large language model can modify a finite state machine from initial configuration $G$ into target configuration $G'$. This evaluation considers various types of modifications that may be necessary to achieve the target configuration, such as adding, deleting, or replacing states, transitions, and conditions. To assess the correspondence between two finite state machines $A$ and $B$, we match all states $S(A)$ and $S(B)$ and all transitions $T(A)$ and $T(B)$ by their unique labels.




\subsection{Data Collection}\label{AA}

 We constructed a dataset based on the source history of the Eindhoven University of Technology's RoboCup@Home team. We selected this repository due to its extensive history, with over 13k commits of Python code, which contains numerous FSMs designed for practical and meaningful tasks, including door opening, human detection, and pick-and-place operations. It also contains contributions from 35 real human programmers and is regularly deployed on physical robots. Because its development spans 9 years, the FSMs have also undergone modifications by humans, allowing us to establish a ground-truth comparison against modifications made by LLMs. 

We analyzed the commit history of files containing FSMs to identify useful data. Specifically, we sampled files, examined them to determine whether they contained state machines, and reviewed their commit history to identify commits containing changes to FSM structures. The structure of a FSM refers to the logical flow of its states, transitions, and conditions. This conceptual framework is distinct from the actual implementation of these three elements, as it focuses on the high-level logic governing the behavior of the FSM rather than the technical details of how they are realized in code or hardware. For each of these commits, we considered the parent (pre-modification) and child (post-modification) versions of the file as potential candidates pair for analysis.

Out of 87 parent-child code pairs reviewed, six contained structural modifications to the FSMs, making them relevant for further investigation. In our study, \textbf{structural changes} refer to modifications that impact the structure of the FSM, such as the addition, removal, or alteration of states, transitions, or conditions. Conversely, non-structural changes include modifications that do not affect the FSM's overall structure, such as renaming states, transitions, or conditions, or changes that affect individual states without altering the broader architecture of the state machine. We used these six pairs for further analysis, taking the parent commit as input (pre-modification FSM code) and the child commit(post-modification FSM code) as the ground truth to see if it represents the same FSM from the code generated by the LLM.

\begin{figure}[t]
    \centering
    \includegraphics[width=9cm]{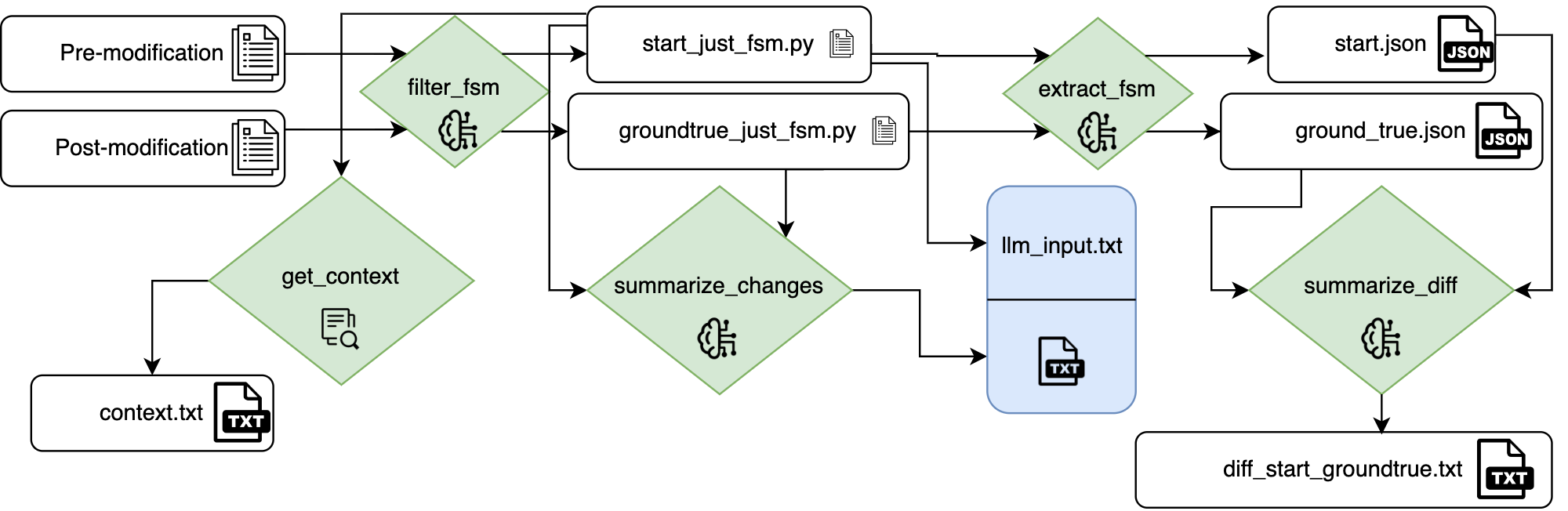}
    \caption{Data processing pipeline consists of multiple stages involving LLM invocations. These stages analyze the dataset by comparing pre-modification and post-modification code snippets to identify differences, extract necessary FSM information, and construct input prompts for the evaluation pipeline.}
    \label{fig:your_image_label}
\end{figure}

\subsection{Data Processing}\label{AA}

We developed a multi-agent system incorporating LLMs as part of our evaluation process. The system leverages zero-shot prompting techniques to enable the agents to perform various tasks as steps of evaluation autonomously. To ensure the correctness and reliability of the system, we manually verified the agents' output. The evaluation tasks, typically time-consuming, benefit significantly from the multi-agent approach, enabling large-scale FSM code evaluation. Although we utilized only six parent-child code pairs in this research, the system has the potential to scale efficiently to evaluate more extensive FSMs. Prior to engaging the LLMs for code modifications, we processed the data to focus on relevant sections. Since our interest lies specifically in the structure of the FSM, we filtered out non-FSM code by recognizing specific patterns, such as the implementation of state machines. In the RoboCup repository, all state machines inherit from the \texttt{smach.StateMachine} class. To capture this inheritance pattern, we employed the following regular expression: 

\begin{verbatim}
class\s+\w+\(smach\.StateMachine\):\s*
(?:\"\"\".*?\"\"\"\s*)?\s*
def\s+__init__.*?
(?=\nclass|\nif\s+__name__|$)
\end{verbatim}

\subsubsection{filter\_fsm Agent}
\vspace{0.5cm}
\hfill\\

We have also developed a version of the \texttt{filter\_fsm} that uses LLM inference. This alternative approach is designed for integration with codebases that do not utilize \textsc{smach} library. The \texttt{filter\_fsm} is tasked with identifying and isolating the key components of the FSM, including state declarations, transitions between states, and the triggering events or conditions. It is important to note that the \texttt{filter\_fsm} disregards the internal implementation details of each state, as well as any unrelated functions or methods that do not contribute to the FSM. The extraction process is designed to handle cases where state transition logic is distributed across multiple classes or methods, ensuring a comprehensive representation of the FSM structure. The resulting output consists solely of the FSM framework, without any embedded code from the individual state methods or irrelevant functions. It is important to note that while we filter the FSM portion of the code during evaluation, this does not imply that the input code for \textit{ChatFSM} must exclusively contain FSMs to yield correct results. As demonstrated later in our analysis, the method remains effective even when additional context is included alongside the FSM. This highlights the inherent capability of LLMs to identify and focus on the FSM components that require modification, regardless of surrounding contextual information.

\subsubsection{extract\_fsm Agent}
\vspace{0.5cm}
\hfill\\

The \texttt{extract\_fsm} is an LLM agent designed to extract FSMs from programming code and convert them into a standardized JSON format. This structured representation simplifies the comparison of FSMs by providing a clear, uniform format that can be easily parsed and analyzed. Furthermore, the JSON output can be utilized to generate visual representations of the FSMs, enabling intuitive comparison and identification of differences between two FSM specifications. The following prompt used for \texttt{extract\_fsm} outlines the task and the expected JSON format:

\begin{quote}
\item Your task is to extract all finite state machines in the code and respond with JSON format. 
\item You should put all FSMs in one JSON file.
\item Your response should strictly follow the format below:

\begin{lstlisting}[language=json]
[
    {"name": "FSM1", 
     "description": "specify the usage of this FSM",
     "initialState": "State1", 
     "states": [
        {"name": "State1", 
         "description": "specify the usage of this state",
         "transitions": [
            {"to": "State2", "outcome": "Event1"},
            {"to": "State3", "outcome": "Event2"}
        ]},
        {"name": "State2", ...},
        {"name": "State3", ...}
    ]},
    {"name": "FSM2", ...}
]
\end{lstlisting}

\end{quote}

By adhering to this format, \texttt{extract\_fsm} ensures consistency in the representation of FSMs, allowing for integration with other tools and workflows that require FSM analysis or visualization. 

\subsubsection{summarize\_changes Agent} 
\vspace{0.5cm}
\hfill\\

The \texttt{summarize\_changes} function is designed to summarize the modifications between old and new versions of a code file in a style resembling a git commit message. Initially, we considered utilizing the original commit messages from the developers of the Robocup repository. However, we found that these messages often lacked detailed information regarding changes made to the FSM. The input for the \texttt{ChatFSM} agent requires the FSM code, followed by specific changes that developers wish to apply. Thus, relying on the original commit messages was not a viable option. The \texttt{summarize\_changes} prompt was structured as follows: 
\begin{quote}
Given two files representing the old and new versions of a codebase, analyze the changes in terms of finite state machines. Identify all the states and transitions that have been added, removed, or modified. Write all these changes in some short commit messages. For example: 'Removed state X, transitioning from Y to Z; Added state A, transitioning from B to C.' Old File:\{file1\} New File:\{file2\}. Make sure to include all changes in terms of finite state machines, all the states and transitions that have been added, removed, or modified. Only output the commit messages, nothing else. Commit messages:\{messages\}
\end{quote}

In our experiment, the code of extracted FSMs from child commit, followed by the output from \texttt{summarize\_changes} (created by concatenating the text data from both files), served as the input for \texttt{ChatFSM} during the evaluation phase. Notably, the agent compares FSMs using the output files of \texttt{filter\_fsm}, rather than comparing extracted JSON files from the parent and child commits. This approach aligns better with our goal of comparing the FSM code directly. An example output of \texttt{summarize\_changes} could be: 

\begin{quote}
    Removed state \texttt{FIND\_PERSON}, transitioning from \texttt{NAVIGATE\_TO\_WAYPOINT} and \texttt{NAVIGATE\_TO\_ROOM} to \texttt{found} or \texttt{not\_found}; 
Added state \texttt{FIND\_PEOPLE}, transitioning from \texttt{NAVIGATE\_TO\_WAYPOINT} and \texttt{NAVIGATE\_TO\_ROOM} to \texttt{found} or \texttt{not\_found}. 
Modified constructor parameters: removed \texttt{area}, \texttt{name}, \texttt{discard\_other\_labels}, and \texttt{found\_person\_designator}; added \texttt{room} and \texttt{found\_people\_designator}.
\end{quote}

\subsubsection{summarize\_diff Agent} 
\vspace{0.5cm}
\hfill\\

We introduce the \texttt{summarize\_diff} function, which processes two extracted JSON files and highlights the differences between them. This agent was prompted as follows:
\begin{quote}
Given two JSON files representing the ground truth and input versions of a finite state machine, analyze the changes in terms of FSM. Identify all the states and transitions that have been added, removed, or modified. Write all these changes in some short messages. Ground truth:\{file1\} Input:\{file2\}. Make sure to include all changes in terms of finite state machines, all the states and transitions that have been added, removed, or modified. Only output the messages, nothing else. Messages:\{messages\}
\end{quote}

An example output of \texttt{summarize\_diff} could be:
\begin{quote}
    \begin{itemize}
    \item State \texttt{SAY\_DETECT\_HANDOVER} added in \texttt{HandoverToHuman}.
    \item Transition from \texttt{MOVE\_HUMAN\_HANDOVER\_JOINT\_GOAL} to \texttt{SAY\_DETECT\_HANDOVER} added in \texttt{HandoverToHuman}.
    \item Transition from \texttt{SAY\_DETECT\_HANDOVER} to \texttt{DETECT\_HANDOVER} added in \texttt{HandoverToHuman}.
    \item State \texttt{SAY\_CLOSE\_NOW\_GRIPPER} added in \texttt{HandoverToHuman}.
    \item Transition from \texttt{DETECT\_HANDOVER} to \texttt{SAY\_CLOSE\_NOW\_GRIPPER} added in \texttt{HandoverToHuman}.
    \item Transition from \texttt{SAY\_CLOSE\_NOW\_GRIPPER} to \texttt{CLOSE\_GRIPPER\_HANDOVER} added in \texttt{HandoverToHuman}.
\end{itemize}
\end{quote}

The use of two comparison agents, \texttt{summarize\_changes} and \texttt{summarize\_diff}, allows for a manual cross-verification of outputs. This dual-agent approach ensures that the LLM accurately detects differences without introducing errors. Moreover, it provides insight into the extent of modifications made by human developers, offering a better understanding of the complexity and difficulty involved in the changes.

\subsubsection{get\_context Agent} 
\vspace{0.5cm}
\hfill\\

We also employ the \texttt{get\_context} agent, which utilizes Retrieval-Augmented Generation (RAG) to retrieve code relevant to changes in the Finite State Machine (FSM) from the entire codebase of the parent commit. This method aims to determine whether including relevant context improves performance as part of our evaluation. Specifically, the retrieved context is appended to the original ChatFSM agent input(\texttt{start\_just\_fsm.py} + summary from \texttt{summarize\_changes}), formatted as follows: \texttt{"Answer the user's questions based on the context below:\textbackslash n\textbackslash n\{context\}"}. This additional step allows us to assess the impact of context incorporation on the model's response accuracy. \texttt{get\_context} has prompt as follows:

\begin{quote}
Finite State Machine(FSM) code follows by Requested Changes \{input\}
Given the above conversation, generate a search query to look up to get all the information relevant to the FSM code and the Requested Changes to the FSM code"
\end{quote}

\subsection{Evaluation Pipeline}\label{AA}

\begin{figure}[t]
    \centering
    \includegraphics[width=9cm]{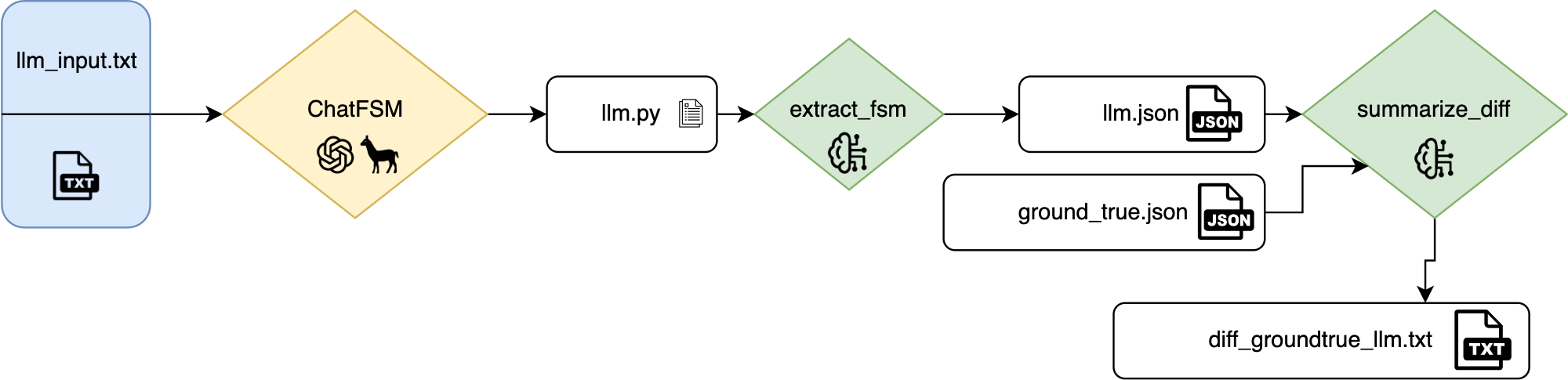}
    \caption{Evaluation pipeline takes the constructed input prompts from the data processing pipeline and summaries the difference from the ground truth. }
    \label{fig:pipeline}
\end{figure}

As shown in Fig. ~\ref{fig:pipeline}, the evaluation begins by providing the input, with or without context, to the LLM. This input is sent to the ChatFSM agent, which outputs a modified Python file or the corresponding code in the language specified by the input. Next, we use the function \texttt{extreact\_fsm} to extract the JSON representation of the FSM. This JSON is then compared with the ground truth, \texttt{ground\_true.json}, using the function \texttt{summarize\_diff}. The comparison result is stored in \texttt{diff\_groundtrue\_llm.txt}. Finally, we analyze this file to evaluate the performance of two selected LLM models: GPT-4o-05-13 and LLaMA-3.1-70b-versatile, utilizing Groq's API.

%% file: sections/5_Results_Discussion.tex
\section{Results}

\subsection{Results}

The results were categorized into three distinct groups. 
\begin{itemize}
    \item \textbf{No Difference}, indicates that the FSM generated by the LLM is identical to the ground truth, with no variations in structure.
    \item \textbf{Small Difference}, includes cases where the FSM differs in non-structural elements, such as the naming of states, transitions, or conditions, but the overall logic and structure remain consistent with the ground truth.
    \item \textbf{Difference}, represents instances where the LLM-generated FSM shows significant structural deviations from the ground truth.
\end{itemize} 

We present the results based on this classification in Table \ref{tab:correctness} and \ref{tab:correctness with context}. 

\input{tables/correctness}
\input{tables/correctness_with_context}

We also observed that the addition of contextual information did not improve performance, as the results remained unchanged. We observed that the retrieved context primarily consists of implementation details for each state in the FSM since it is what we asked it to do in the prompt. Consequently, this additional information does not contribute meaningfully to modifying the FSM structure, as it does not offer information beyond the predefined state transitions. 

Both models exhibit a small difference(\texttt{small\_diff}) in pair 3, specifically in the transition condition for the state \texttt{DECIDE\_NAVIGATE\_STATE}, where the difference was: \textit{"- Transition condition changed in state \texttt{DECIDE\_NAVIGATE\_STATE}: 'none' to 'not\_found'."} This divergence arises because the language model (LLM) was unaware of the exact conditions for \texttt{DECIDE\_NAVIGATE\_STATE} and hallucinated the condition name \texttt{'not\_found'}. Additionally, the retrieval-augmented generation (RAG) model failed to provide the implementation of the \texttt{\_DecideNavigateState} class as part of the context. After manually incorporating this class into the context, we reran the evaluation for this pair, and both models produced correct results with no difference. This suggests that providing more comprehensive context can improve the accuracy of implementation details, beyond just the structure of the FSM. 

ChatFSM still successfully modified the structure of FSMs across all six parent-child file pairs. The effectiveness of these modifications can be assessed by examining the complexity of changes made to the FSMs by the developers. Below is a summary of the FSM changes for each pair:

\textbf{Pair 1}: 1 state replaced, 2 transitions reconnected to the replaced state as source. The robot is changed from looking for a person to looking for a crowd of people.
\begin{figure}[h]
    \centering
    \includegraphics[width=0.175\textwidth]{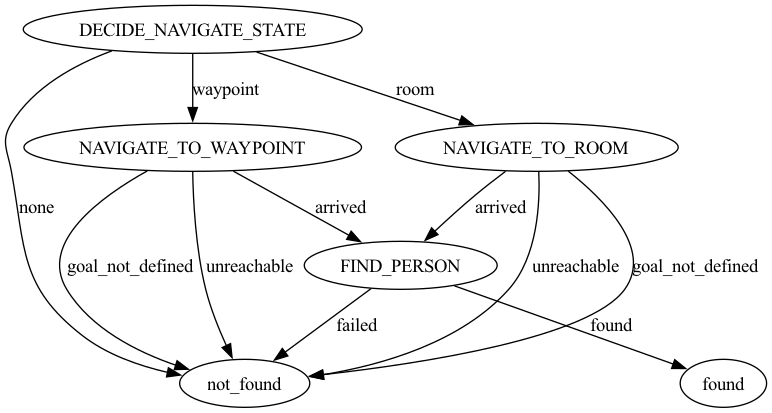} \hspace{0.5cm} 
    $\rightarrow$ 
    \hspace{0.5cm} 
    \includegraphics[width=0.175\textwidth]{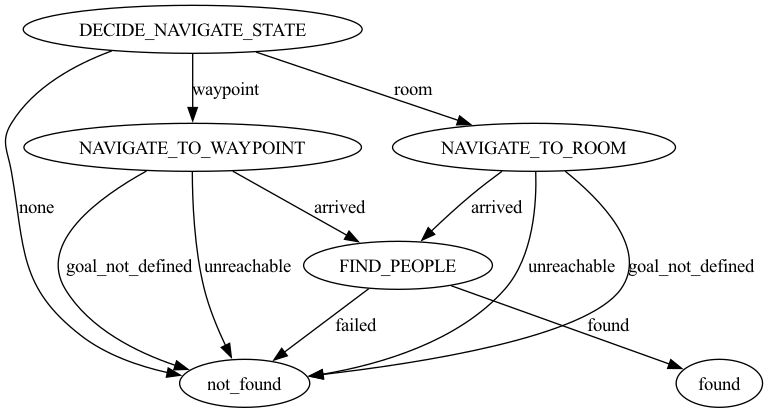}
    \caption{Pair 1}
    \label{fig:pair 1}
\end{figure}

\textbf{Pair 2}: 5 states removed, 1 state added, 10 transitions removed, 6 transitions added.
The robot's behavior is changed from sequential gripper and arm control actions to a more streamlined handover detection process. Instead of explicitly commanding the gripper to open or close, the robot now detects the handover event directly, reducing verbal feedback when the gripper is open and closed. 

\begin{figure}[h]
    \centering
    \includegraphics[width=0.175\textwidth]{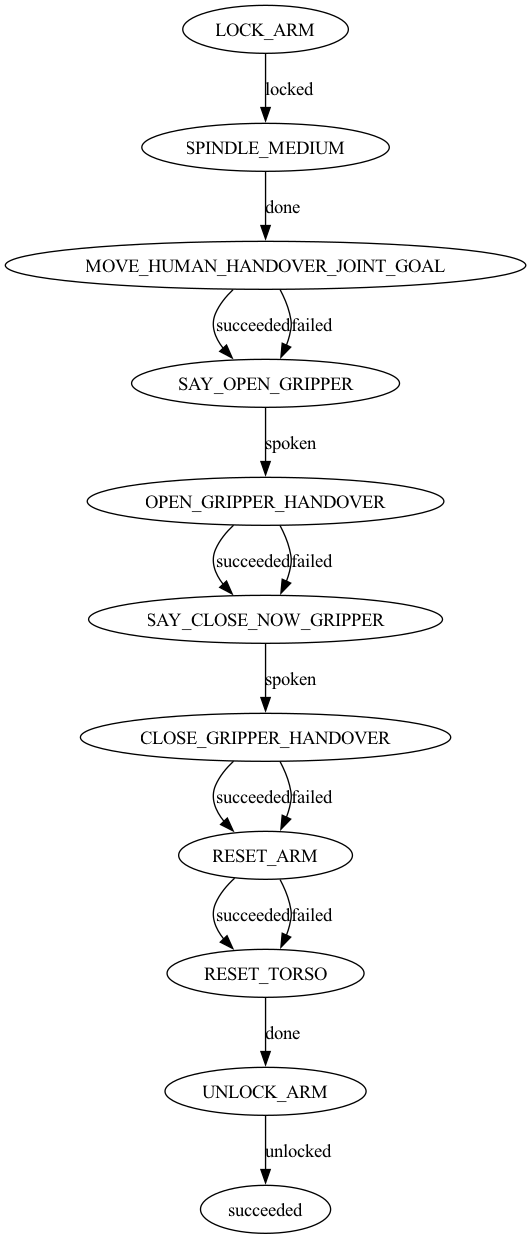} \hspace{0.5cm} 
    $\rightarrow$ 
    \hspace{0.5cm} 
    \includegraphics[width=0.175\textwidth]{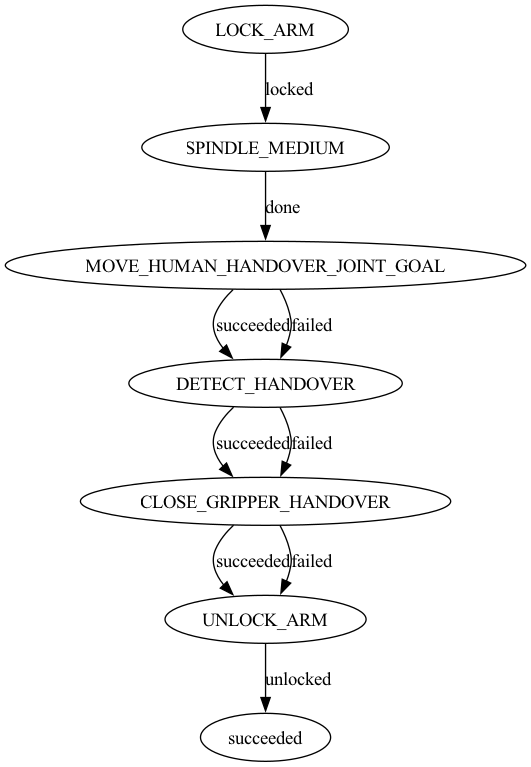}
    \caption{Pair 2}
    \label{fig:pair 2}
\end{figure}

\textbf{Pair 3}: 1 state removed, 3 states added, 9 transitions added, 4 transitions removed, 1 transition rewired to the one of the added states as source.
The robot's behavior is changed from navigating directly to a predefined area to first deciding between navigating to a specific waypoint or room before searching for a person. In the new state machine, the robot now enters a decision-making phase, where it evaluates whether to proceed to a waypoint or room. Once it reaches the waypoint or room, the robot then searches for a person. 

\begin{figure}[h]
    \centering
    \includegraphics[width=0.175\textwidth]{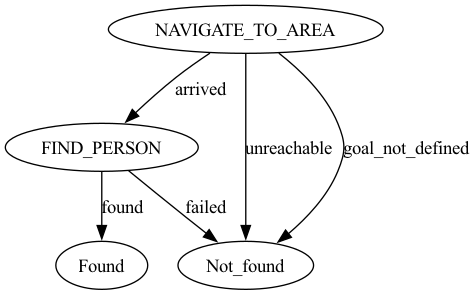} \hspace{0.5cm} 
    $\rightarrow$ 
    \hspace{0.5cm} 
    \includegraphics[width=0.175\textwidth]{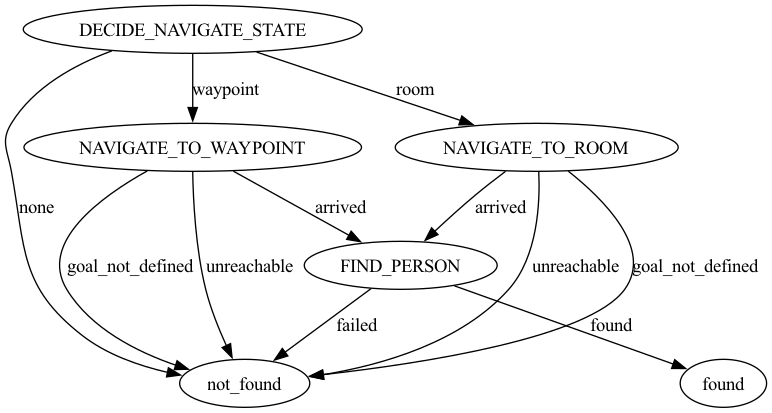}
    \caption{Pair 3}
    \label{fig:pair 3}
\end{figure}
        
\textbf{Pair 4}: 1 state added, 2 transitions added, 2 conditions added.
The robot's behavior is changed from ending the gripper handover process after success or failure to resetting its arm position before proceeding to unlock the arm. Previously, upon either a successful or failed handover, the robot would directly transition to a final state, but with these modifications, the robot now actively resets its arm position after every handover attempt before continuing to unlock the arm, ensuring more consistent readiness for subsequent actions.

\begin{figure}[h]
    \centering
    \includegraphics[width=0.175\textwidth]{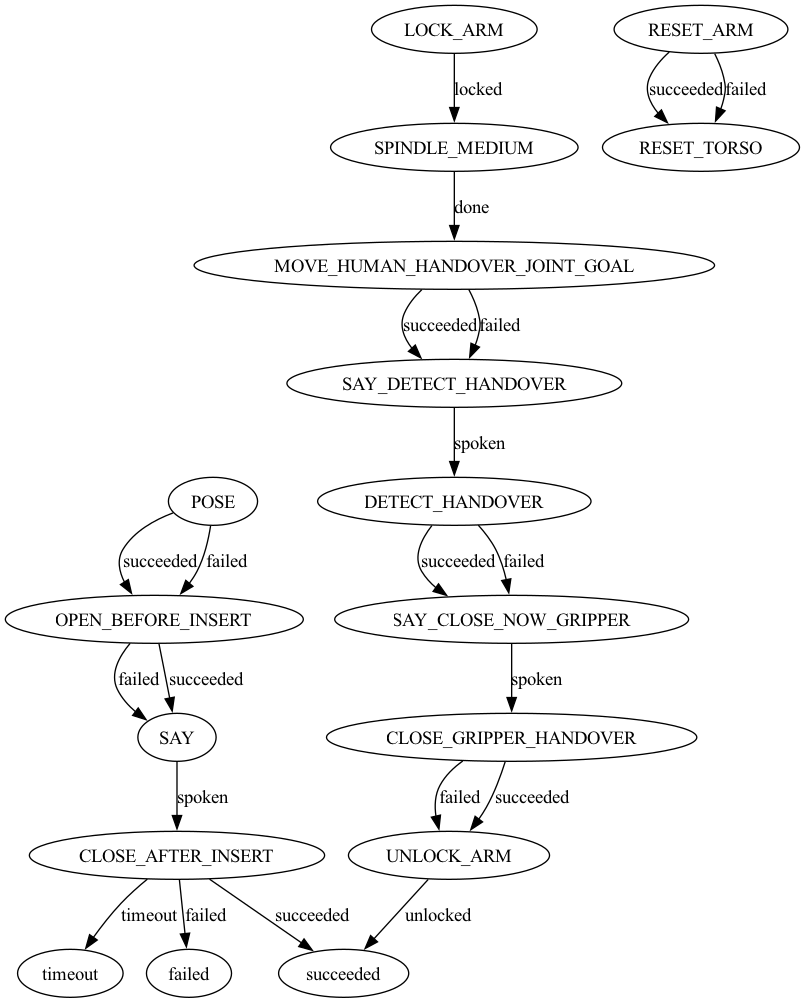} \hspace{0.5cm} 
    $\rightarrow$ 
    \hspace{0.5cm} 
    \includegraphics[width=0.175\textwidth]{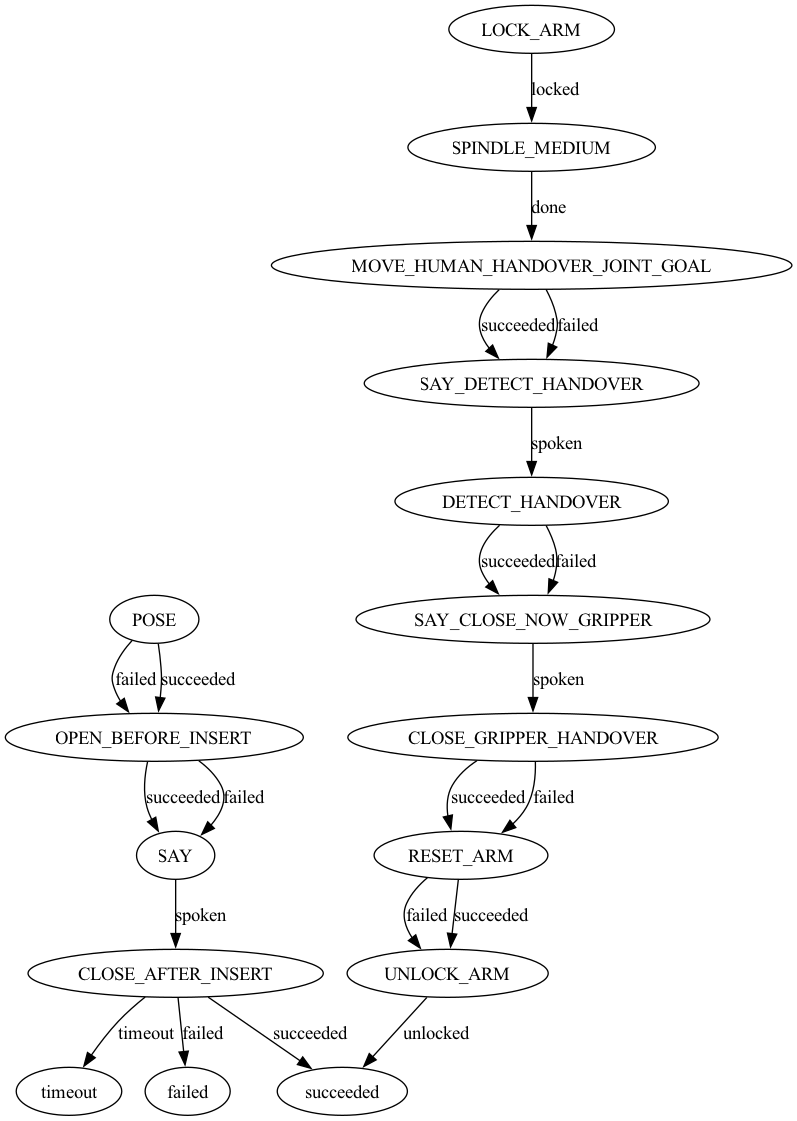}
    \caption{Pair 4}
    \label{fig:pair 4}
\end{figure}
        
\textbf{Pair 5}: 2 states added, 4 transitions added.
The robot is changed from directly detecting a handover after moving to the human handover joint goal to first announcing the detection step. And the robot now vocalizes or indicates that it is about to detect the handover. Additionally, after detecting the handover, the robot signals that it is about to close its gripper. 

\begin{figure}[h]
    \centering
    \includegraphics[width=0.175\textwidth]{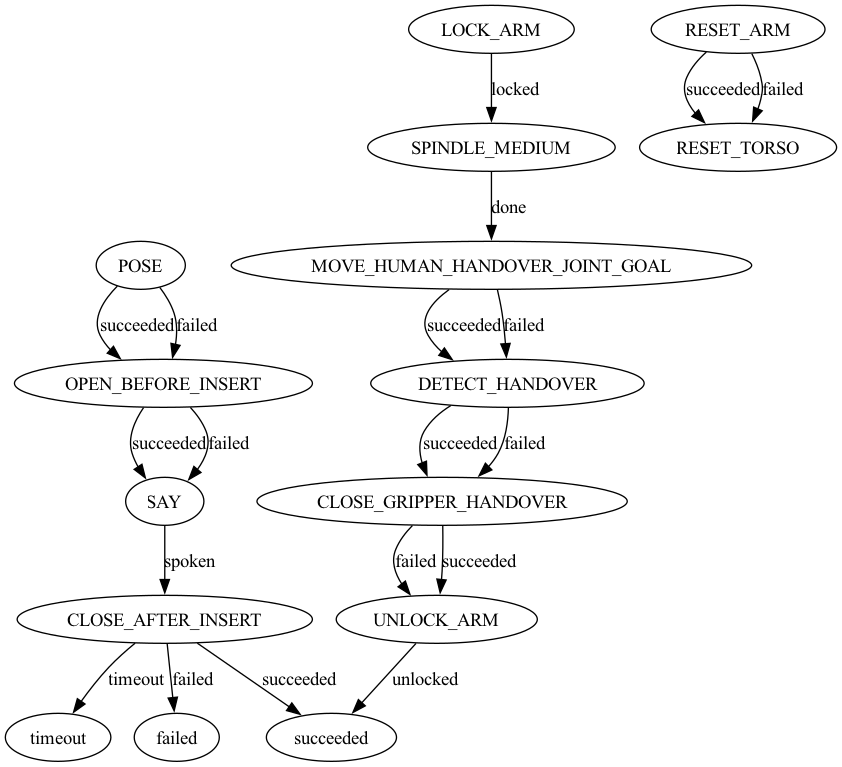} \hspace{0.5cm} 
    $\rightarrow$ 
    \hspace{0.5cm} 
    \includegraphics[width=0.175\textwidth]{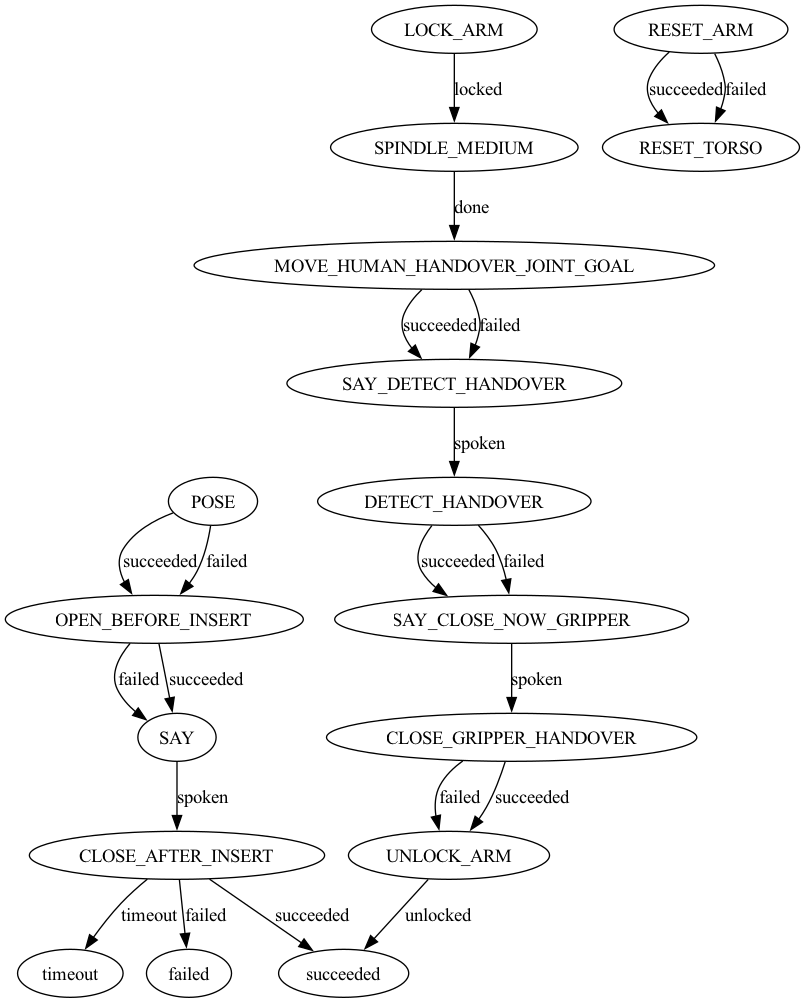}
    \caption{Pair 5}
    \label{fig:pair 5}
\end{figure}
        
\textbf{Pair 6}: 2 states removed, 3 transitions removed, 2 conditions modified.
 Previously, the robot would handle cases where head goal position were canceled before moving on. Now, the robot bypasses these checks, leading to a logic where it simply proceeds based on whether the learning was successful or not, without needing to account for head goal position cancellations

 \begin{figure}[h]
    \centering
    \includegraphics[width=0.175\textwidth]{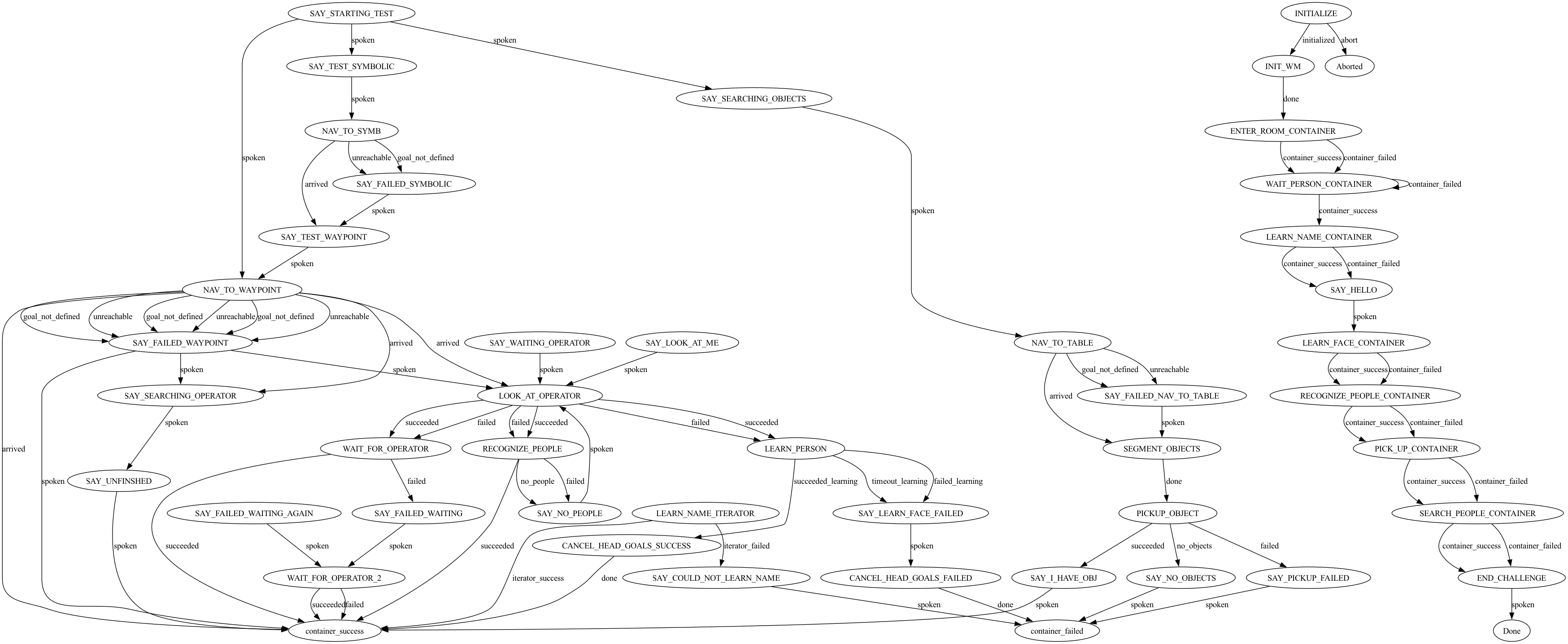} \hspace{0.5cm} 
    $\rightarrow$ 
    \hspace{0.5cm} 
    \includegraphics[width=0.175\textwidth]{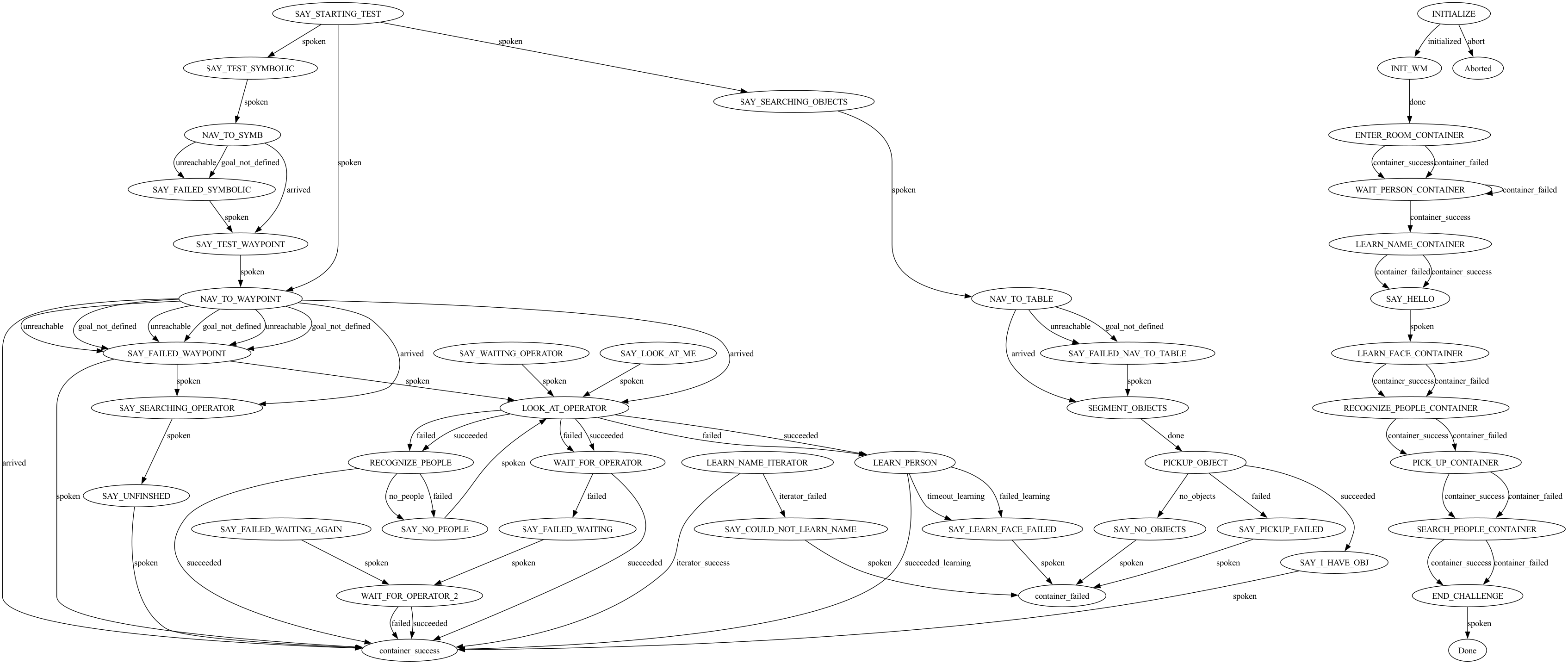}
    \caption{Pair 6}
    \label{fig:pair 6}
\end{figure}

The FSMs can be modified and updated rapidly with the help of LLMs compared manual modifications, as shown in Table~\ref{tab:processing_times}. 

\input{tables/time}

\subsection{Limitations}

FSM generation is another crucial aspect of FSM automation, which remains unexplored in our study. While our results demonstrate the LLM's capacity to add states to an existing FSM based on user-provided natural language instructions, we did not showcase its ability to generate the corresponding implementation code for each state. This represents an important area for future work, as the ability to automatically generate both the high-level FSM structure and the functional state implementations would significantly enhance the practical applicability of LLMs in robotic software development.

The evaluation of our approach was based on a relatively small number of samples, which may limit the generalizability of our findings. A larger, more diverse set of FSM modification tasks and examples would provide greater confidence in the robustness and versatility of the LLM's performance across various contexts and domains. The limited sample size further constrains the ability to fully assess how well the LLM can handle a wide range of modifications, from simple state additions to complex restructuring of state machines.

Moreover, we only focused on a single FSM framework, derived from the version control history of the Eindhoven University of Technology's RoboCup@Home team. While this framework is representative of a real-world robotic application, it does not capture the diversity of FSM frameworks and coding conventions used across different domains and robotics projects. The LLM's capabilities may vary significantly when applied to other FSM frameworks with different architectures, libraries, and conventions. Future work should aim to evaluate the performance of LLMs across multiple FSM frameworks and coding styles to better understand their generalizability and adaptability.

Additionally, the codebase we utilized for our evaluation presented challenges, particularly the lack of natural and descriptive textual context. For instance, the commit messages in the repository were often terse, lacking detailed descriptions of the changes being made, the intent behind them, or the high-level logic of the FSM modifications. This presented a difficulty for the LLM, as it relies on contextual information to accurately generate and modify code. In practice, the ability of an LLM to perform FSM modification and generation effectively depends on having access to rich contextual information, including meaningful commit messages, comments, and function names. Addressing this issue would involve curating datasets with more informative commit histories or using synthetic examples where the FSM modifications are accompanied by clear and explicit context.

In summary, while our work demonstrates the potential of LLMs in FSM modification, there remain several limitations and open questions. These include the need for larger and more varied datasets, evaluation across multiple FSM frameworks, and improvements to contextual information available in codebases to fully realize the capabilities of LLMs in FSM automation. Future research in these areas would be instrumental in advancing the application of LLMs for FSM generation and manipulation in real-world robotic systems.

%% file: tables/correctness.tex
\begin{table}[t]
\begin{center}
\begin{tabular}{|c|c|c|c|}
\hline
\textbf{Model} & \multicolumn{3}{|c|}{\textbf{Difference Categories}} \\
\cline{2-4} 
 & \textbf{\textit{No Diff}} & \textbf{\textit{Small Diff}} & \textbf{\textit{Diff}} \\
\hline
gpt-4o-05-13 & 5 & 1 & 0 \\
\hline
llama3.1-70b & 5 & 1 & 0 \\
\hline
\end{tabular}
\end{center}
\caption{ChatFSM Correctness}
\label{tab:correctness}
\end{table}

%% file: tables/correctness_with_context.tex
\begin{table}[t]
\begin{center}
\begin{tabular}{|c|c|c|c|}
\hline
\textbf{Model} & \multicolumn{3}{|c|}{\textbf{Difference Categories}} \\
\cline{2-4} 
 & \textbf{\textit{No Diff}} & \textbf{\textit{Small Diff}} & \textbf{\textit{Diff}} \\
\hline
gpt-4o-05-13 & 5 & 1 & 0 \\
\hline
llama3.1-70b & 5 & 1 & 0 \\
\hline
\end{tabular}
\end{center}
\caption{ChatFSM Correctness With Context}
\label{tab:correctness with context}
\end{table}

%% file: tables/time.tex
\begin{table}[h]
\centering
\begin{tabular}{|l|c|}
\hline
\textbf{Model} & \textbf{Processing Time (s)} \\ \hline
GPT-4o-2024-05-13 & 59.4 \\ \hline
Llama-3.1-70b-Versatile & 3.7 \\ \hline
Human & 164 \\ \hline
\end{tabular}
\caption{Average Processing Times for Different LLMs}
\label{tab:processing_times}
\end{table}

%% file: sections/6_Conclusion.tex
\section{Conclusion}

The results of our research demonstrate that recent advancements in LLMs are indeed capable of accurately modifying the structure of FSMs across a diverse set of file pairs. Our findings reveal that, regardless of the complexity of changes—ranging from minor state modifications to extensive rewiring and transition restructuring—the LLMs consistently and correctly executed the required transformations.

This proficiency in executing a wide range of FSM modifications showcases the potential for LLMs to automate code refactoring and structural changes, streamlining the process of updating robot behavior logic. By enabling developers to specify FSM modifications through natural language interactions, LLMs offer a promising tool for reducing the time and effort involved in managing FSMs, which are integral to both software development and robotics applications. 

Our research, therefore, provides a positive answer to the question posed in this paper’s title: "Can Large Language Models Help Developers with Robotic Finite State Machine Modification?" The results suggest that LLMs hold significant potential for facilitating FSM manipulation, making them a valuable asset in the toolkit of robotic developers seeking to efficiently modify and enhance robotic behavior through high-level language inputs.

%% file: sections/7_Acknowledgement.tex
\section*{Acknowledgment}
\ifthenelse{\boolean{anonymized}}{
    ChatGPT was utilized to assist with the proofreading of the text and generation of LaTeX syntax in this work.
}{
    ChatGPT was utilized to assist with the proofreading of the text and generation of LaTeX syntax in this work.
}